\documentclass{ecai} 

\usepackage{latexsym}
\usepackage{amssymb}
\usepackage{amsmath}
\usepackage{amsthm}
\usepackage{booktabs}
\usepackage{enumitem}
\usepackage{graphicx}
\usepackage{color}

\usepackage{algorithm}
\usepackage{algpseudocode}

\usepackage{multirow}%
\usepackage[title]{appendix}%
\usepackage{xcolor}%
\usepackage{tabularx}
\usepackage{makecell}
\usepackage{ragged2e}
\usepackage{enumitem} 

\usepackage{dsfont}
\usepackage{siunitx}
\usepackage{float}
\newtheorem{definition}{Definition}

\newcommand{\BibTeX}{B\kern-.05em{\sc i\kern-.025em b}\kern-.08em\TeX}

\newcommand{\R}{\mathbb{R}}

\newcommand{\A}{\mathcal{A}}

\newcommand{\abs}[1]{{\left|#1\right|}}
\DeclareMathOperator*{\argmax}{arg\,max}
\DeclareMathOperator*{\argmin}{arg\,min}

\newcommand{\conc}{{\textproc{conc}}}
\newcommand{\chr}{\textproc{chr}}
\newcommand{\repr}{\textproc{repr}}
\newcommand{\VS}{\text{VS}}
\newcolumntype{Y}{>{\Centering\arraybackslash}X}
\begin{document}

%%%%%%%%%%%%%%%%%%%%%%%%%%%%%%%%%%%%%%%%%%%%%%%%%%%%%%%%%%%%%%%%%%%%%%%%

\begin{frontmatter}

%%% Use this command to specify your submission number.
%%% In doubleblind mode, it will be printed on the first page.

\paperid{6755} 

%%% Use this command to specify the title of your paper.

\title{Learning the Value Systems of Societies from Preferences}

%%% Use this combinations of commands to specify all authors of your 
%%% paper. Use \fnms{} and \snm{} to indicate everyone's first names 
%%% and surname. This will help the publisher with indexing the 
%%% proceedings. Please use a reasonable approximation in case your 
%%% name does not neatly split into "first names" and "surname".
%%% Specifying your ORCID digital identifier is optional. 
%%% Use the \thanks{} command to indicate one or more corresponding 
%%% authors and their email address(es). If so desired, you can specify
%%% author contributions using the \footnote{} command.

\author[A]{\fnms{Andrés}~\snm{Holgado-Sánchez}\orcid{0000-0001-8853-1022}\thanks{Corresponding Author. Email: andres.holgado@urjc.es.}\footnote{Equal contribution.}}
\author[A]{\fnms{Holger}~\snm{Billhardt}\orcid{0000-0001-8298-4178}\footnotemark}
\author[A]{\fnms{Sascha}~\snm{Ossowski}\orcid{0000-0003-2483-9508}\footnotemark} 
\author[B]{\fnms{Sara}~\snm{Degli-Esposti}\orcid{0000-0003-0616-8974}}
\address[A]{CETINIA, Universidad Rey Juan Carlos, 28933 Madrid, Spain}
\address[B]{CSIC, Consejo Superior de Investigaciones Científicas, 28006 Madrid, Spain}
%%% Use this environment to include an abstract of your paper.

\begin{abstract}
Aligning AI systems with human values and the value-based preferences of various stakeholders (their value systems) is key in ethical AI. In \emph{value-aware} AI systems, decision-making draws upon explicit computational representations of individual values (groundings) and their aggregation into value systems. As these are notoriously difficult to elicit and calibrate manually, value learning approaches aim to automatically derive computational models of an agent's values and value system from demonstrations of human behaviour. 
Nonetheless, social science and humanities literature suggest that it is more adequate to conceive the value system of a society as a set of value systems of different groups, rather than as the simple aggregation of individual value systems. Accordingly, here we formalize the problem of learning the value systems of societies and propose a method to address it based on heuristic deep clustering. The method learns socially shared value groundings and a set of diverse value systems representing a given society by observing qualitative value-based preferences from a sample of agents. We evaluate the proposal in a use case with real data about travelling decisions.

\end{abstract}

\end{frontmatter}

%%%%%%%%%%%%%%%%%%%%%%%%%%%%%%%%%%%%%%%%%%%%%%%%%%%%%%%%%%%%%%%%%%%%%%%%

\section{Introduction}\label{sec:introduction}
Value alignment in AI~\cite{Russell2022alignmentDefinition} deals with the problem of aligning the objectives and functioning of AI systems with human values. Defining human values and value-based preferences (or \textit{value systems}) is a challenging task because values vary across time and cultures. In addition, at the time of acting, human preferences may be incomplete due to incommensurable values and context-specificity. Nevertheless, as humans, we expect software agents to be locally coherent and to develop some ability of normative reasoning \cite{zhi2024beyond}. %As a result, relevant works in the literature propose an implicit way to achieve value alignment through deep learning~\cite{christiano2023deeprlpreferences}, for example by imitating human preferences~\cite{Watson2024}, or through cooperation~\cite{hadfieldCIRL}. 
Recently, authors argue that truly value-aligned AI systems must be able to explicitly reason about the consequences of their behaviour (or the ones of their acquaintances) based on specific human values~\cite{valueengineeringAutonomous2023}, allowing their adaptation to the value systems of different stakeholders~\cite{andres2024vecompPaper}. This explicitness of value alignment (aka \emph{value awareness}) has been approached through classical multi-criteria decision making setups~\cite{AaronAgreementTOPSIS,Karanik2024}, in reinforcement learning (RL)~\cite{manel2022ethical}, or via semantic representations such as taxonomies~\cite{Osman2024} or ontologies~\cite{de2022basicValueNet}. 

%However, despite the obvious advantages over implicit alignment representations, \textit{value awareness} approaches present the challenge of instantiating these components correctly. If done through manual design, the process is prone to misspecification~\cite{Sumers2022InstructionsAndDescriptions}. As a pertinent solution, the problem of value learning~\cite{Soares2018ValueLearningProblem} has been introduced as learning these computational models through automatic processes from demonstrations of (value-aligned) behaviour. 
Value awareness approaches face the challenge of correctly instantiating their models. As manual design is prone to misspecification~\cite{Sumers2022InstructionsAndDescriptions}, value learning~\cite{Soares2018ValueLearningProblem} suggests to induce  them  automatically from demonstrations of value-aligned behaviour. To this respect, the most common concern is value identification~\cite{Liscio2022Axies2}, which refers to the problem of identifying stakeholders' value preferences or the set of values specific to a certain context (from texts, stakeholder opinions, etc.), together with value system estimation~\cite{Siebert2022liscio}. 

%Value identification and value system estimation do not necessarily require finding a proper computational representation of individual values, though. This makes the approaches less useful for value-aware systems, that would require computational value representations for value reasoning, with only few applicable related works (that do not model specific human values) on logic programming~\cite{Anderson2018} and reinforcement learning~\cite{Leike2018ScalableAA,Peschl2022LearnPreferencesFromExperts}.

%For example, doctors abide by bioethical principles and comply with the same medical protocols~\cite{towardsAwarenessMedicalField2024}. Even though humans are different and hold different value systems based on their values~\cite{Serramia2020valuesystem}, it is undeniable that they grow up in a society full of values and norms that shape their preferences. 

%Thus, the fundamental problem is how to represent the value system of a society as a whole, considering all the nuances given by individual decisions. 
Values are intrinsically social and are shared among groups of humans (societies)~\cite{Osman2024}. Given the value systems of a set of agents in a certain group, value aggregation~\cite{leraleri2024aggregation,AaronAgreementTOPSIS} consists of estimating the value system that better represents their values. Still, learning methods for this task demand heavy human moderation, namely, that the agents give a numerical estimation of the alignment of every possible decision in the world with all values considered. Also, according to~\cite{leraleri2024aggregation}, value systems are \textit{pluralistic}, and thus, considering a single value system in a society can misrepresent value system diversity.

In this paper, we propose a \textit{social value system learning} method that extends previous work~\cite{andres2024vecompPaper} and aims at representing the value systems of a society of agents by observing diverse agent choices. Our contribution is three-fold. Firstly, we put forward a formal definition of the ``value system of a society'' that includes (a) a socially-agreed value \textit{grounding} model to computationally represent value alignment with a given set of values, and (b) a clustering of agents in terms of the similarity of their value preferences, stated in terms of the previous grounding. We also enunciate desirable properties for such a social value system, namely the \textit{grounding coherency, representativeness and conciseness}. Secondly, we propose a formulation of the problem of learning the value system of a given society based on a structured optimization of the previous properties, tackled through observing stated pairwise comparisons between alternatives by different agents based on values and individual preferences. Finally, we present a joint preference learning and clustering algorithm based on MaxMin-RLHF~\cite{pmlr-v235-chakraborty24b} that provides an approximate solution to the proposed problem.
To evaluate our contributions, we consider a real-world use case in train route choice modelling~\cite{apollochoicepublication}. Apart from demonstrating the capability of the algorithm to solve the enunciated problem, we evaluate whether the learned value systems reflect stated human intentions, such as choosing trips for shopping or business.

%The paper is organized as follows. In Section~\ref{sec:stateofart}, we overview related work. In Section~\ref{sec:background}, the necessary notions for modelling the value systems of single agents from previous work are presented.  Section~\ref{sec:proposal} describes 
%our first and second contributions, namely, 
%the proposed definition of the value system of a society, its desirable properties, and the formulation of the learning problem. Section~\ref{sec:algorithms} explains our algorithmic solution of the formulated problem. In Section~\ref{sec:evaluation}, we evaluate and discuss our contributions in the mentioned use case and Section~\ref{sec:conclusion} presents concluding remarks, limitations and suggestions for future work.

The paper is organized as follows. Section~\ref{sec:stateofart} overviews related work. Section~\ref{sec:background} presents needed notions for modelling value systems of single agents from previous work. Section~\ref{sec:proposal} describes 
%our first and second contributions, namely, 
the proposed definition of the value system of a society, its desirable properties, and the formulation of the learning problem. Section~\ref{sec:algorithms} explains our algorithmic solution. In Section~\ref{sec:evaluation}, we evaluate and discuss our contributions in the mentioned use case and Section~\ref{sec:conclusion} presents conclusions, limitations and future work suggestions.

\section{Related work}\label{sec:stateofart}

The novel field of Value Awareness Engineering (VAE)~\cite{valueengineeringAutonomous2023} claims that, to achieve value-aligned behaviour in real-world domains, agents must be able to reason with and about values. For this purpose, they need to explicitly model value meaning or alignment in a computational manner; an approach called by some authors \textit{operationalizing} values~\cite{Shahin2022OperationalizingvaluesSurveySoftwareengineering}. Most of these models are based on mathematical functions that measure the degree by which agent-based or system-based states~\cite{montes2022synthesis}, actions/decisions~\cite{Karanik2024}, or both~\cite{manel2022ethical} are effectively aligned with values, or their meaning \textit{grounded} in a particular domain. On top of explicit models of values, and in order to provide solutions for value-based decision-making (DM) and negotiation, some authors model also human value systems, either quantitatively (assuming a set of value weights~\cite{negotiationOfNormsvalues2021,Karanik2024,andres2024vecompPaper}) or qualitatively (via order relations between values~\cite{Serramia2020valuesystem}).

For value awareness, it is necessary to operationalise human values~\cite{Shahin2022OperationalizingvaluesSurveySoftwareengineering}. \emph{Value identification}~\cite{Liscio2022Axies2} refers to the process of identifying the set of values relevant to stakeholders. %It is a pivotal element of the normative alignment problem in \cite{Gabriel2020}.
It is typically addressed by social surveys and experiments~\cite{schwartz2005schwartz}, or through data-driven methods such as value classification in texts~\cite{QiuZhaoLiLuPengGaoZhu2022ValueNetDataset}. A second task is  \emph{value system estimation}~\cite{Siebert2022liscio} which infers the value systems of individuals.

A key aspect often overlooked is that value awareness requires grounding value meanings and preferences in specific domains for computational use. When done automatically, this process is known as value learning~\cite{Soares2018ValueLearningProblem}. Some approaches relevant to value learning, though not explicitly modelling values, include GenEth\cite{Anderson2018}, which learns ethical principles as \textit{prima facie} duties; or Pesch et al.\cite{Peschl2022LearnPreferencesFromExperts}, who use inverse reinforcement learning to learn norm-compliant rewards and trajectory preferences reflecting various value systems. An example of explicit value modelling is~\cite{userStudyLearningValues}, which learns alignment models from user studies in healthcare. In our previous work~\cite{andres2024vecompPaper}, we addressed value learning from behaviour traces.
     
%\subsection{Learning alignment through preferences}\label{sec:stateoftheartIRL}

Leike et al.~\cite{Leike2018ScalableAA} claim that value alignment can be achieved through careful reward modelling. As such, some authors are inclined to learning human alignment through preference-based~\cite{christiano2023deeprlpreferences} or inverse reinforcement learning~\cite{ng2000algorithms} by learning reward models (or directly, aligned policies/behaviours) explaining human demonstrations. Approaches that make use of preference learning in value alignment are limited. Loreggia et al.~\cite{Loreggia2019MetricPreferencesAlignment} learn a quantitative metric from a given partial order between options with neural networks. Some approaches go beyond by jointly learning multiple goals and preferences in multiobjective RL from demonstrations~\cite{Mu2024Preference-basedModeling,Kishikawa2022Multi-ObjectiveEstimation}. In our previous work, we applied a similar idea to achieve explicit value-alignment with various stakeholders~\cite{andres2024vecompPaper}.

%\subsection{Value Aggregation}

Outside of computer sciences, authors consider the value system alignment problem~\cite{Macedo2013ValueSystemAlignmentICollaborativeEnvironments}, i.e., finding the degree of alignment between the value systems of agents in a society, considering the compatibility of the values of different agents. Under cases where a certain degree of compatibility exists, the problem of value aggregation~\cite{Siebert2022liscio} in societies of agents is a natural problem to address, i.e., finding a value system that represents a group of agents through negotiation or social choice. For instance, \cite{AaronAgreementTOPSIS} utilizes TOPSIS to reach conflict-free or agreed value systems, while \cite{leraleri2024aggregation} relies on l$p$-regression to reach a consensus-based value system aligned with ethical principles varying from maximum utility to maximum fairness.  Although not based on values, fine-tuning LLMs through human feedback~\cite{Watson2024} implicitly aggregated the values and preferences of many people. Introducing ways for enhancing group representation through, e.g., social choice, has been identified as a key future line of work~\cite{pmlr-v235-chakraborty24b} together with preference personalization~\cite{li2024personalizedlanguagemodelingpersonalized}. The existing works in this area, while relevant, fail at characterizing the diverse user preferences in terms of explicit goals or values.

%%%%%%%%%%%%%%%%%%%%%%%%%%%%%%%%%%%%%%%%%%%%%%%%%%%%%%%%%%%%%%%%%%%%%%%%

\section{Representing values and value systems}\label{sec:background}
We set out from a set of $m$ values $V = \{v_1, ..., v_m\}$, where each value $v_i$ is conceived as a label for a particular value. When \textit{grounded} in a specific domain, a value label acquires a particular meaning. We model this meaning through the notion of the \textit{alignment} of a set of entities in the domain with the value. Depending on the domain, the set of entities might be the set of alternatives in a classical DM stance, or, rather, the outcomes that these alternatives provoke. For example, in route choice analysis, the entities of study could be the paths or routes that the agent can traverse; whereas in government policy making, the set of entities could consist of the outcomes that these policies provoke in society. We assume humans can elicit the value alignment of entities qualitatively. Formally, we assume a notion of value alignment based on a preference relation between entities.
\begin{definition}[Value Alignment]\label{def:value-alignment-preference}
     The alignment of a set of entities $E$ with a value $v_i$ is represented by a weak order $\preccurlyeq_{v_i}$ over $E$, where $e\preccurlyeq_{v_i} e'$ means that $e'$ is at least as aligned with value $v_i$ as $e$.
\end{definition}

Following similar works in the area~\cite{Serramia2018,montes2022synthesis}, we claim humans have some inherent value alignment function for each value $v_i$ (difficult to elicit or unknown), $\A_{v_i}$, that represents the qualitative alignment relation $\preccurlyeq_{v_i}$ such that for all $e, e' \in E $:
    $$ e \preccurlyeq_{v_i} e' \iff \A_{v_i}(e) \leq \A_{v_i}(e') $$

To specify the semantics of a set of values, we define the notion of \textit{grounding}.
%We then define \textit{grounding} as the set 
\begin{definition}[Grounding]\label{def:grounding}
    A \textbf{grounding} of the set of values $V$ is a set of weak orders $\preccurlyeq_{V} = \left\{\preccurlyeq_{v_i}\right\}_{i=1}^m$. Given the respective alignment functions, a \textbf{grounding function} for $V$ is: $G_V=\left(\A_{v_1}, \dots, \A_{v_m}\right)$. 
\end{definition}

Agents build their individual value systems on top of a grounding, considering alignment preferences within a certain domain. 

\begin{definition}[Value system]\label{def:value-system}
Let $V$ be a finite set of values, and let $\preccurlyeq_{V}$ be a grounding for $V$. The \textbf{value system} of an agent $j$ is a weak order $\preccurlyeq^j_{V}$ over $E$ derived from the grounding $\preccurlyeq_{V}$. If $e\preccurlyeq^j_{V} e'$, we say that $e$ is equally or more aligned than $e'$ with the $j$'s value system. 
\end{definition}

%We can make the derivation of an agent's value system from grounded values (which is based on the importance that different values have for that agent) explicit using grounding functions. 
Given a grounding function, the value system of an agent can be represented employing a value system function.

%\begin{definition}[Value System Function]\label{def:value-system-alignment-function}
%Let $j$ be an agent with a value system $\preccurlyeq^j_{V}$ and grounding function $G_V$. The function $\A_{f_j,{G_V}} : E \to \R$ with $\A_{f_j,{G_V}}(e) = f_j(\A_{v_1}(e), \dots, \A_{v_m}(e))$ is a \textbf{value system function} for agent $j$ if it represents $\preccurlyeq^j_{V}$ over $E$, i.e., for all $e,e' \in E$:
%$$\A_{f_j,{G_V}}(e) \leq \A_{f_j,{G_V}}(e') \iff e \preccurlyeq^j_{V} e'$$ where $f_j: {\R}^m \to \R$ is an aggregation function that combines the value alignment with respect to each value. 
%\end{definition} 

\begin{definition}[Value System Function]\label{def:value-system-alignment-function}
Let $j$ be an agent with a value system $\preccurlyeq^j_{V}$ and grounding function $G_V$. The function $\A_{f_j,{G_V}} : E \to \R$ with $\A_{f_j,{G_V}}(e) = f_j(\A_{v_1}(e), \dots, \A_{v_m}(e))$ is a \textbf{value system function} for agent $j$ if it represents $\preccurlyeq^j_{V}$ over $E$, i.e.:
$$\forall e,e' \in E: \ \A_{f_j,{G_V}}(e) \leq \A_{f_j,{G_V}}(e') \iff e \preccurlyeq^j_{V} e'$$ where $f_j: {\R}^m \to \R$ is an aggregation function that combines the value alignment with respect to each value. 
\end{definition} 

To keep value system functions simple and interpretable, we restrict them to linear scalarization functions, frequently used in multi-objective decision-making~\cite{linearscalarizedMOMDP2013}. We represent $f_j$ through a set of positive \textit{value system weights} $W_j = (w^{v_1}_j\dots, w^{v_m}_j)$ with $\sum_{i=1}^m w_j^{v_i} = 1$. Thus, $\A_{f_j,{G_V}}(e) = \sum_{i=1}w_j^{v_i}\A_{v_i}(e)= W_j\cdot {G_V}^T(e)$.

%\textcolor{blue}{The previous weak orders also define strict preference relations. If $e \preccurlyeq e'$ and $\neg (e\succcurlyeq e')$, we write $e\prec e'$ to assert that $e$ is strictly better aligned than $e$ (with a value system or value). If we have both $e \preccurlyeq e'$ and $e \succcurlyeq e'$ ($e\approx e'$), we say they are equivalently aligned.}

\section{Representing the value system of a society}\label{sec:proposal}

As outlined in Section~\ref{sec:introduction}, values are inherently socially relevant notions~\cite{schwartz2005schwartz,Osman2024}, and different agents hold different value systems~\cite{leraleri2024aggregation}, which makes relevant the problem of describing the value system(s) of a society. In the following, we assume that for a given application domain, there is a society of agents $J$, where each agent has an individual value system $\preccurlyeq_{V}^j$ based on a certain grounding $\preccurlyeq_{V}$.

Regarding the grounding, we consider that agents can potentially have varied perspectives on the meaning of values. However, within human societies and certain application domains there exist typically a  \textit{socially-agreed} grounding~\cite{andres2024vecompPaper}, i.e., there is a consensus on how value alignment is understood. Stating this social agreement is a way of recognizing that morality is universal, yet culturally variable \cite{haidt2007new}. All humans have moral intuitions, which are fast processes in which an evaluative feeling of good-bad or like-dislike (about the actions or character of a person) appears in consciousness and is later followed by moral reasoning. Simmel, Durkheim, Parsons, and other authors used the word \emph{socialization} to refer to the mechanism that enables social reproduction, that is, the reproduction of value systems over time. The idea of social grounding reflects this tradition of studies and serves to acknowledge that we live in a social milieu full of values; we decide which of these values to endorse or abandon. %From a technical perspective, this assumption is not limited to a domain. For instance, beneficence, non-maleficence, autonomy, and justice are ethical principles shared in the healthcare domain~\cite{clinicalValues2020}. %where as part defined in medical protocols that all hospitals agree upon

%The previous assumption has important technical implications. If we can represent a socially-agreed grounding, we can build value systems that are much more meaningful, in the sense that they are more true to the prior value understandings. Also, different value systems of different agents can be more easily compared with one another if we use the same groundings of values as reference.

We represent a socially-agreed grounding using a grounding function $G_V$. %The aggregation of groundings from different agents into a consensus is performed in value aggregation works with certain success~\cite{leraleri2024aggregation}.
To quantitatively assess its coherence for individual agent groundings, we rely on evaluating how well it represents these.
%the following form of agent demonstrations. 
We consider datasets $D_{v_i}^j$ for each agent $j$ and value $v_i$, containing pairs of entities on which the agents state their value alignment preferences. Each entry $(e, e', y)\in D_{v_i}^j$ captures whether agent $j$ believes $e$ is more aligned with $v_i$ than $e'$ ($y=1$), less aligned ($y=0$), or equally aligned ($y=0.5$). We denote the full \textit{grounding dataset} as $D_V = \{D_{v_i}^j|j\in J,v_i\in V\}$. Note that we do not assume agents rank the same entities and not all possible pairs of them. %which is an unfeasible assumption in infinite entity spaces and requires heavy moderation in general.

We then define a quantitative transformation representing the relative alignment difference of two entities from a candidate alignment function $\A_{v_i}$. Following previous work~\cite{andres2024vecompPaper}, we employ the Bradley-Terry model, frequently used for preference modelling from pairwise comparisons datasets~\cite{christiano2023deeprlpreferences} for learning reward models (Eq.~(\ref{eq:comparisonquantity})).
\begin{equation}\label{eq:comparisonquantity}
p(e,e'|\A_{v_i}) = \frac{\exp{\A_{v_i}(e)}}{\exp{\A_{v_i}(e)}+ \exp{\A_{v_i}(e')}}
\end{equation}
Notice that, effectively, $p(e,e'|\A_{v_i}) = 0.5$ only if $\A_{v_i}(e) = \A_{v_i}(e)$ and it tends to $1$ or $0$ if their difference in alignment is increasingly strict. With this model, we can formally define the coherence of a value alignment function $\A_{v_i}$ with the alignment preferences of a set of agents manifested through the previous datasets.

%For each agent in the studied society $j \in J$ we observe a dataset $\mathcal{D}_j = \{(e_i,e'_i,y^j_i)\}_{i\in I_j}$, where $y^j_i \in [0,1]$ captures $j$'s relative preference for entity $e_i$ over $e'_i$ according to $j$'s value system. We consider also another dataset from domain experts (or from the very same agents), namely, $\mathcal{D}_V = \{(e_i,e'_i,y^{v_1}_i,\dots, y^{v_m}_i)\}_{i\in I_V}$ where $y^{v_l}_i \in [0,1]$ for all $l\in \{1,\dots,m\}$. Though they are defined generally as quantitative measures in $[0,1]$, both $y_l^{v_l}$ and $y_i^{j}$ may represent qualitative preferences if they are restricted to the values $\{0,1,0.5\}$, to indicate that $e'_i$ is preferred over $e_i$, the inverse relation, or indifference, respectively. 

\begin{definition}[Coherence of a value alignment function/grounding]

Let $J$ be a society of agents. The coherence of a value alignment function $\mathcal{A}_{v_i}$ for value ${v_i}$ over a dataset of agent-based alignment preferences $D_{v_i} = \{D_{v_i}^j|j\in J\}$, is given by:

\begin{align*}
    \chr_{D_{v_i}}(\mathcal{A}_{v_i}) =1-\frac{1}{\abs{J}} \sum_{j\in J}\frac{1}{\abs{D_{v_i}^j}}\sum_{(e,e',y)\in D_{{v_i}}^j} \delta(p(e,e'|\A_{v_i}), y)
\end{align*}

%$$
%ch(e,e'|\A_v,y) = \begin{cases}
%1 \ \text{if} \begin{cases}
    %\abs{p(e,e'|\A_v)-y} < 0.5\ \text{ if }\ y \in \{0,1\}\\
    %\abs{p(e,e'|\A_v)-y} \sim 0\ \text{ if }\ y =0.5
    %\end{cases}\\
    %0 \ \text{otherwise}
%\end{cases},
    %$$
\[\text{where: }
\delta(p, q) =
\begin{cases}
0 & \text{if } \left(p, q = \frac{1}{2}\right) \lor \left(p,q > \frac{1}{2}\right) \lor \left(p,q < \frac{1}{2}\right) \\
1 & \text{otherwise}
\end{cases}
\]
The coherence of a grounding function $G_V = (\A_{v_1}, \dots, \A_{v_m})$ is the average over $V$: $\chr_{D_V}(G_V) = \frac{1}{m}\sum_{i=1}^m \chr_{D_{v_i}}(\mathcal{A}_{v_i})$

\end{definition}
The function $\delta(p,q)$ measures the disagreement between $p$ and $q$ assuming they represent an alignment preference over a certain pair of alternatives using the Bradley Terry model (Eq.~(\ref{eq:comparisonquantity})). It is $0$ if both $p$ and $q$ agree with respect to the alignment preference and $1$ if not. %i.e. if both are bigger than, equal, or smaller than $0.5$, as it would mean the alignment functions they assumedly originate from assign more, equal or less alignment to the first option than to the second, respectively.
Here, we use it to see if the preference model obtained from $\A_{v_i}$ disagrees with the stated alignment preferences ($y$) for pairs $e,e'$.

The socially-agreed assumption implies that a grounding function with high coherence should exist. A grounding function with coherence 1, means that it fully aligns with all stated agents' preferences. %If the agents are rational, the socially-agreed assumption implies that such a function exists, or at least one with very high coherency allowing maybe for minor discrepancies that don't significantly affect the overall value semantics.

We now define the value system of a society. Naturally, there can be more discrepancies in the value preferences between stakeholders~\cite{leraleri2024aggregation}, and in principle each agent might have its own, different value system. Nevertheless, assuming that people growing up in the same social milieu have their value system influenced by culture \cite{grenfell2014pierre}, we can expect regularities in the value systems of agents in the same social groups. Given this, and recalling our social grounding assumption, we propose representing the value system of a society as the composition of a (socially-agreed) grounding together with a set of value systems, tentatively representing different groups of agents determined through a certain assignment function.

\begin{definition}[Value system of a society]\label{def:society-value-system}
Let $\preccurlyeq_{V}$ be a grounding for a set of values $V$ over entities $E$ and let $J$ be a society of agents.

A \textbf{value system of the society} $J$ , $\VS^{J,L,\beta}_{V}$, is a family of $\abs{J} \geq L\geq 1$ value systems $\left\{\preccurlyeq^l_{V} \middle| l \in \{1, \dots, L\}\right\}$  derived from $G_V$  over $E$, together with an assignment function $\beta : J \to \{1,\dots, L\}$ that assigns each agent to one of the $L$ value systems. We define the group of agents assigned to the $l$-th value system ($\preccurlyeq^l_{V}$) by $C_l = \{j \in J | \beta(j) = l\}$ and call it the $l$-th \textit{cluster} of the society.
\end{definition}

%&As per Definition~\ref{def:society-value-system}, a value system of the society can vary from being a single value system shared by the population (with $L=1$), or be composed by a different one for each of the agents (when $L=\abs{J}$), with all intermediate groupings possible. The first option is the most \textit{concise}, while the latter is the most \textit{representative} of agent preferences. We aim to evaluate the quality of a society's value system for all values of $L$ and assignments $\beta$, balancing the trade-off between these two goals. To model this trade-off mathematically, we draw a parallel with clustering analysis~\cite{Kettenring2006clusteringinterintra}, which involves grouping elements into clusters by balancing intra-cluster and inter-cluster distances/similarities. The notion of \textit{representativeness} is a parallel with intra-cluster similarity, a measure for the degree that every agent is represented by its assigned value system/cluster. \textit{Conciseness} can be regarded as a parallel with inter-cluster distance, measuring the differences between clusters/value systems in terms of how they differ in the entity preferences they entail: maximizing this would yield value systems that are different and aggregate agents into opposable groups. In the following we define the necessary terms to define these two notions formally.
According to Definition~\ref{def:society-value-system}, the number of value systems in a society ($L$) can range between 1 and $\abs{J}$. The former is the most concise, and the latter is the most representative regarding individual preferences. Our goal is evaluating the quality of value systems across different values of $L$ and assignments $\beta$, balancing these two goals. To formalize this trade-off, we draw upon an analogy with cluster analysis. Representativeness parallels intra-cluster similarity --how well each agent is represented by its assigned value system--. Conciseness parallels inter-cluster distance --how distinct the value systems are in terms of the preferences they induce--. In the following we define these two concepts formally.

Like for value alignment, for each agent $j$ we assume we have access to a dataset $D_{\VS}^j$ with samples of stated preferences between entities with respect to $j$'s value system.
 Its entries are of the form $(e, e',y)$, where $y \in \{0, 0.5, 1\}$ indicates whether $j$ strictly prefers $e$ over $e'$ ($y = 1$), strictly prefers $e'$ over $e$ ($y = 0$) or is indifferent between both options ($y = 0.5$) (always according to $j$'s value system). We write $D_{VS}^J$ for the union of all agent-dependent datasets.  %We can partition $D_{\VS}^j$ by value if needed, yielding ${D_v^{j,\VS} \mid v \in V}$, and aggregate across agents to obtain $D_v^{\VS} = \bigcup_{j \in J} D_v^{j,\VS}$ for each value or $D_{\VS} = \bigcup_{j \in J} D_{\VS}^j$ for the full dataset.

In a society, we estimate the $l$-th value system with a value system function $\A_{W_l,G_V} = W_l\cdot {G_V}^T$ where $G_V$ is the (socially-agreed) grounding function, and $W_l \in [0,1]^m$ represent the value system weights. We define with this, the discordance of a value system function with the value system of an agent $j$ enacted through $D_{\VS}^j$ by:
\begin{equation}\label{eq:discordance}
    d_{D_{\VS}^j}\left(\A_{W_l,G_V}\right) = \frac{1}{\abs{D_{\VS}^j}}\sum_{(e,e',y)\in D_{\VS}^j}\delta(p(e,e'|\A_{W_l,G_V}),y)
\end{equation}
%\chr\left(e_i,e_i'\middle|\A_{W_l,G_V},y_i^j\right)$$ 
%where
%$$ch(e,e'|\preccurlyeq,y) = \begin{cases}
    %1 \text{ if }\{\mathds{1}\left(e \succcurlyeq e'\right)-%\mathds{1}\left(e \preccurlyeq e'\right)=2y-1\}\\
   % 0 \text{ otherwise}
%\end{cases}$$

%The discordance metric in Definition~\ref{def:representativeness} allows to define our parallel notion with intra-cluster similarity, the \textit{representativeness} of a value system. Representativeness is a score that measures the degree by which a social value system \textit{represents} the value systems of the individuals. We define it generally in terms of abstract value systems, the adaptation for datasets $D_{\VS}$ and value system functions is straightforward.

%Using the discordance, we define representativeness of a value system of a society to measure how well the individual agents' preferences are correctly represented by the value systems of the clusters the agents are assigned to.

Using the discordance, we define representativeness of a value system of a society as the degree by which each agents' preferences are represented by the value systems the agents are assigned to.

\begin{definition}[Representativeness of the value system of a society]\label{def:representativeness}
    Let $\VS^{J,L,\beta}_{V}$ be a value system of the society $J$ and let $D_{\VS}^j$ be a preference dataset for each agent $j\in J$.
    %Let the set of all value systems of the individual agents $\left\{\preccurlyeq^j_{V} \middle| j \in J\right\}$. Let, for each agent $j\in J$, some set $P_j \subset E\times E$. Let $P = \bigcup_{j\in J} P_j$ and $P_l = \bigcup_{j\in \beta^{-1}(l)}P_j$ (the set of pairs compared by all the agents assigned to the $l$-th cluster). 
    %The representativeness of the $l$-th cluster of the society, represented through the value system function $W_l$ and $G_V(\cdot)$ over the dataset $D_{\VS}^J$ is:
    The representativeness of $\VS^{J,L,\beta}_{V}$, represented by the value system weights $W=\{W_l\}_{l=1}^L$  and the grounding function $G_V$ over the dataset $D_{\VS}^J$ is:
    %$$\repr_{P_l}\left(\preccurlyeq_{G_V}^l|\ C_l\right) = \frac{1}{\abs{C_l}}\sum_{j\in C_l} \left(1 - d_{P_j}\left(\preccurlyeq_{G_V}^{l}, \preccurlyeq_{G^j_V}^{j}\right)\right)$$

    %$$\repr_{D_{\VS}^l}\left(C_l|W_l,G_V\right) = 1-\frac{1}{\abs{C_l}}\sum_{j\in C_l} d_{D_{\VS}^j}\left(\A_{W_l,G_V}\right)$$

    $$\repr_{D_{\VS}^J}\left(\VS^{J,L,\beta}_{V}\middle|{W,G_V}\right) = 1-\frac{1}{\abs{J}}\sum_{j\in J}d_{D_{\VS}^j}\left(\A_{W_{\beta(j)},G_V}\right)$$

    %OLD: $$\repr_P\left(\VS^{J,L,\beta}_{G_V}\right) = \min_{l=1,\dots,L} \repr_{P_l}\left(\preccurlyeq_{G_V}^l|\ C_l\right)$$

\end{definition}

 Representativeness is the main goal for a social value system, for it promotes a configuration of value systems that better represent the individuals assigned to them. It is bounded in $[0,1]$, with $1$ indicating a maximum and $0$ a minimum level of representation.  %It can be regarded as a qualitative version of the consensus-based goal in ${l}$p value aggregation~\cite{leraleri2024aggregation} for many aggregated value systems}.
 
 Maximizing representativeness does not prohibit having two or more individual value systems producing similar preferences. Our second clustering-inspired measure, \textit{conciseness} (proxy for inter-cluster distances) should alleviate this problem. We define it as the minimum discordance between each pair of value systems of the society, considering the comparisons made by each agent.
 
 %The aggregation over all the value systems is the minimum over each cluster representativeness, which ensures representation fairness among the different groups. 

%A \textit{maximally} representative social value system is one that, if it replaced the individual value systems of the agents in the society, their preferences would still be represented to the highest possible extent. Note that (in our abstraction) a trivial maximally representative value system that may achieve a maximum representativeness always exist: the one formed by the $L=\abs{J}$ original value systems of the agents. Of course this situation is not desirable at all as it would provide no useful information about the values of the society. Besides, having two or more individual value systems having the same preferences (redundancy) is not penalized in this score. As explained before, we need to consider a trade off with the second clustering goal, that is, maximizing somehow the inter-cluster distances, the discordance between the value systems conforming the society. To model this we employ a notion of value system \textit{conciseness}.

\begin{definition}[Conciseness of the value system of a society]\label{def:conciseness}
Let $\VS^{J,L,\beta}_V$ be a social value system.
%and let $\{A_{W_l,G_V}\}_{l=1}^L$ denote the value system functions for the $L$ clusters. 
The conciseness of  $\VS^{J,L,\beta}_V$ 
represented through the value system weights $W=\{W_l\}_{l=1}^L$  and the grounding function $G_V$ over the dataset $D_{\VS}^J$ is defined by:
%Let $VS_{G_V}^{J,L,\beta}$ a value system of the society $J$ with $L$ value systems and assignment function $\beta$. Let $P = \bigcup_{j\in J}P_j$. The conciseness of $VS_{\{J, G_V,L\}}$ over $P$ is a distance:

    %$$\conc_P(VS_{G_V}^{J,L,\beta} = \frac{1}{\binom{L}{2}}\sum_{1\leq l<l'\leq L} d_{P_l\cup P_{l'}}(\preccurlyeq_{G_V}^{l}, \preccurlyeq_{G_V}^{l'}) $$
 %   $$\conc_P\left(VS_{G_V}^{J,L,\beta}\right) = \min_{1\leq l<l'\leq L} d_{P}(\preccurlyeq_{G_V}^{l}, \preccurlyeq_{G_V}^{l'}) $$
 $$\conc_{D_{\VS}^J}\left(VS_{V}^{J,L,\beta}\middle|{W,G_V}\right) = \min_{\substack{l\neq l'\\ 
\abs{C_l}>0\\|C_{l'}|>0}} d_{D_{\VS}^J}(\A_{W_l,G_V}, \A_{W_{l'},G_V}), $$ 
\begin{equation}\label{eq:discordance-inter}
    d_{D_{\VS}^J}(\A, \A') = \frac{1}{\abs{J}} \sum_{j\in J}\sum_{(e,e',\_)\in D_{\VS}^j} \frac{\delta(p(e,e'|\A),p(e,e'|\A'))}{\abs{D_{\VS}^j}}
\end{equation}

\end{definition}

Conciseness is based on the minimum discordance between any pair of value system functions, i.e. based on counting (and averaging) the disagreement ($\delta$) between the respective preference models over the dataset (Eq.~(\ref{eq:discordance-inter})).  The closer the conciseness is to $1$, the higher the separation between the value systems in terms of the preferences they induce. A conciseness of $0$ indicates that there are at least two value systems that are equivalent in their induced preferences. Maximizing conciseness amplifies diversity in the found value systems, which tends to decrease the number of used clusters. When $L = 1$, conciseness is not defined: in this case, a good social value system can simply be described by its representativeness. Conciseness promotes the variety and uniqueness of value systems in the society.

We are now in position to define the \textit{social value system learning problem} addressed in this paper. It consists of the following bi-level optimization problem. Given a society $J$ and datasets 
     $D_{\VS}^J$ and $D_V$, find a value system $VS_V^{J,L^*,\beta^*}$ represented by the value system weights $W^* = \{W_l\}_{l=1}^{L^*}$ and grounding function $G^*_V$ such that:
    \[
(W^*, L^*, \beta^*) \in \argmax_{W, L, \beta} \ \frac{\conc_{D_{\VS}^J}\left(VS_V^{J,L,\beta} \mid W, G_V^*\right)}{1 - \repr_{D_{\VS}^J}\left(VS_J^{J,L,\beta} \mid W, G_V^*\right)}
\]
\[
\text{and subject to} \quad G_V^* \in \argmax_{G_V} \ \chr_{D_V}(G_V)
\]
    %where $G_V = (\A_{v_1}, \dots, \A_{v_m})$ and $G^j_V = (\A^j_{v_1}, \dots, \A^j_{v_m})$ .

%\begin{definition}[Maximally-aligned value system of a society]\label{def:maximally-aligned-vs-problem}
 %    A maximally-aligned value system of $J$ over datasets $D_{\VS}^J$ and $D_V$ is a value system $VS_V^{J,L^*,\beta^*}$ represented by the value system weights $W^* = \{W_l\}_{l=1}^{L^*}$ and grounding function $G^*_V$ that satisfies:
     
    %$$L^*, \beta^* \in \argmax_{L, \beta} \frac{conc_P\left(VS_{V}^{J,L,\beta}\right)}{\max_{j\in J} d_{P_j}(\preccurlyeq^{\beta(j)}_{G^*_V},\preccurlyeq^{j}_{G^j_V})}$$
    
  %  \[
%(W^*, L^*, \beta^*) \in \argmax_{W, L, \beta} \ \frac{\conc_{D_{\VS}^J}\left(VS_V^{J,L,\beta} \mid W, G_V^*\right)}{1 - \repr_{D_{\VS}^J}\left(VS_J^{J,L,\beta} \mid W, G_V^*\right)}
%\]
%\[
%\text{subject to} \quad G_V^* \in \argmax_{G_V} \ \chr_{D_V}(G_V)
%\]
    %where $G_V = (\A_{v_1}, \dots, \A_{v_m})$ and $G^j_V = (\A^j_{v_1}, \dots, \A^j_{v_m})$ .
%\end{definition}
This formulation promotes a social value system to scope for two goals in a hierarchical manner, i.e., maximizing a trade-off between conciseness and representativeness, but only with value systems built on maximally coherent groundings. The trade-off is managed through an adaptation of the Dunn Index~\cite{Dunn01011974}, which originally comprises the division of the minimum inter-cluster distance and the maximum intra-cluster distance. In our case, the numerator corresponds to the conciseness, and the denominator to the negated representativeness. %The latter does not correspond exactly to what the Dunn Index expects, but in our case, we do not need the strict claim that all clusters fairly capture the preferences of their assigned agents, rather, that each agent is, in general, well represented by the value system it is assigned to. 
In the following we use ``Dunn Index'' to refer to our conciseness-coherence ratio. In the supplementary material, we discuss alternative clustering metrics to the Dunn Index.

%In our case, the numerator is the same as in the original index, the way conciseness is defined as a minimum between cluster distances. The denominator is softened to minimizing the representativeness, i.e. the \textit{average} intra-cluster similarity, instead of the original maximum, but the intentionality is the same. As explained before, other aggregation methods can be proposed: depending on the application domain, if misrepresenting preferences in certain clusters is not acceptable, we should consider the minimum cluster representativeness, while if this not an issue, the current averaging formulation suffices. 

%Definition~\ref{def:maximally-aligned-vs-problem} also applies with $L=1$ with a subtle modification. When a candidate to maximally-aligned value system of a society comprises a single value system ($L=1$), the first goal is reduced to maximum representativeness, by making the conciseness of this candidate equal to the best found-so-far clustering with $L > 1$. 

The bi-level optimization setup is needed instead of first estimating a coherent grounding and then trying to learn a social value system. We show this in the supplementary material.

Solving this bi-level problem ensures learning good value alignment models as a prerequisite to final preference elicitation. This offers advantages over pure deep RLHF approaches, which typically mix preferences with goals and lack intermediate representations~\cite{christiano2023deeprlpreferences}. The bi-level structure also promotes alignment models (groundings) compatible with linear weights to represent diverse value systems —improving on prior similar works that assume such weights should exist from fixed value representations~\cite{manel2022ethical,andres2024vecompPaper}. Additionally, the clustering score favours a minimal number of diverse, representative value systems, each defined by interpretable weights. Efficient solutions to this problem thus improve over other personalized preference learning methods that overlook concise clusters~\cite{pmlr-v235-chakraborty24b} or rely on non-interpretable user embeddings~\cite{li2024personalizedlanguagemodelingpersonalized}.

%Solving this bi-level problem guarantees learning good models of value alignment as a prerrequisite prior to final preference elicitation. We argue this poses advantages over similar preference elicitation techniques such as pure deep RLHF approaches, which do not adapt to intermediate representations normally, mixing preferences with goals~\cite{christiano2023deeprlpreferences,dpoLLM2023}. Also, due to the bi-level structure, our problem formulation implicitly promotes value alignment models that support the usage of simple linear weights over them to model the variety of observed value system preferences. This is relevant, as it overcomes the limitation of previous value alignment works that assume or learn reward functions related to values assuming that linear weights on these rewards will actually model real behaviours/preferences~\cite{manel2022ethical,andres2024vecompPaper}. Furthermore, the clustering score should tend to minimize the number and maximize the variety of the obtained value systems while maintaining their representativeness. Each cluster is defined by interpretable value system weights. Thus, methods to efficiently solve the proposed problem should provide an advantage over other personalized human preference learning works that neglect the conciseness of the representation~\cite{pmlr-v235-chakraborty24b,park2024rlhfheterogeneousfeedbackpersonalization}, or rely on embeddings~\cite{li2024personalizedlanguagemodelingpersonalized} and latent variables~\cite{sriyash2024variationalRLHF} to implicitly model different user preferences.

\section{Algorithm}\label{sec:algorithms}
%\textbf{Option 1}: Learn a value system of each agent (estimating $\preccurlyeq^{j}_{G^j_V}$), subject to grounding correctness. Then clustering.

%\textbf{Option 2}: Learn all on the fly with a bi-level optimization approach 

%\textbf{Clustering approach: Hierarchical / EM approach}
%Hierarchical: can try each L very easily.
%EM approach
%\textbf{Clustering static or dynamic: }
%\textbf{Static}: clustering estimating on the fly $\preccurlyeq_{G_V}^{\beta(j)}$ as the formal aggregation of the rewards of all agents inside a cluster. Then train networks to recreate the resulting aggregation. More complexity, time consumption, but much clearer debugging process... Not clear that an aggregation is good. Only applicable in with option 1, option 2 would be overkill

%\textbf{Dynamic}: \textbf{EM approach with exploration}. Simpler, faster probably.

To approximate a solution of the stated learning problem, we propose a combination of two algorithms: Algorithm~\ref{alg:algorithm1} to find a social value system through clustering and Algorithm~\ref{alg:algorithm2} to manage exploration of new solutions and the improvement of existing ones.

We propose a clustering approach based on deep learning. A key parameter of the algorithm is a \textit{maximum} number of clusters $L_{max}$. The algorithm approximates a solution for the \textit{social value system learning problem} with no more than $L_{max}$ clusters.

%With regard to the clustering algorithm, we propose an approach based on deep learning. A key parameter of the algorithm is a \textit{maximum} number of clusters $L_{max}$. The algorithm approximates a solution for the \textit{social value system learning problem} with no more than $L_{max}$ clusters.

We consider two kinds of neural networks. First, the network $G^\theta_V : \Phi \to \R^m$ with parameters $\theta_V$, that represents a socially-agreed grounding function $G_V$ by observing features of the entities residing in a certain space $\Phi$. We also consider $L_{max}$ neural networks each consisting of a linear layer given by certain value system weights $W_l^\omega$ that are parametrized with $\omega \in \R^m$. The weights are calculated from the parameters $\omega$ through a \textit{softmax} calculation $W^\omega_l=(w_j^{v_1}, \dots, w_j^{v_m}) = \frac{\exp \omega}{\sum \exp(\omega)}$. This ensures that they are positive and normalized. In the algorithm, given an assignment $\beta$ with $L$ used clusters, we only consider the value system weights/networks with populated clusters. At every moment, we set $W_j \equiv W^\omega_{\beta({j})}$, estimating each agent's value system $\preccurlyeq_V^j$ with the value system function of the corresponding cluster $\A_l^{\omega,\theta} \triangleq \A_{W^\omega_l,G^\theta_V} = W^\omega_{\beta(j)}\cdot \left(G_V^\theta\right)^T$.

The algorithm is based on EM (Expectation-Maximization) clustering, mimicking~\cite{pmlr-v235-chakraborty24b}. There, the approach was used to learn a clustering of agents in terms of their preferences regarding pairs of options. To do so, it performs several times a cycle of two steps. In the first step, the algorithm assigns each agent to the cluster (a preference model) that represents its preferences better (E-Step, Lines 3-6). In the second step (M-Step, Lines 7-13), the preference model of each cluster is trained to better fit the preferences of the assigned agents. 

The M-step from~\cite{pmlr-v235-chakraborty24b} consists on fitting a reward model $R^\theta(e)$ minimizing a cross-entropy-like loss on the training data: 

$$\mathcal{L}\left(e,e',y\middle|R^\theta\right) = -y\log(p(e,e'|R^\theta)-(1-y)\log(p(e,e'|R^\theta))$$ 

In our case, we have to fit $2$ groups of reward models (alignment functions). The first group is one model per value, i.e., the grounding function $G_V^\theta=\left(\A_{v_1}^\theta,\dots, \A_{v_m}^\theta\right)$; the second is composed by up to $L_{max}$ value system functions, that depend on the weights $W^\omega_l$, $l=1,\dots, L_{max}$ and the grounding models $G_V^\theta$. %Moreover, these two reward models need to be fitted in a hierarchical manner (bi-level optimization). 
Each group of models depend on different datasets, which suggests two groups of loss functions, one based on the value system dataset $\mathcal{L}_{\VS}(D_{\VS}^J|\beta)$, at Eq.~(\ref{eq:loss_vs}), and another consisting of one loss per value of the grounding dataset $\mathcal{L}_{V}(D_V)$, at Eq.~(\ref{eq:loss_gr}).

\begin{align}\label{eq:loss_vs}
    &\mathcal{L}_{\VS}(D_{\VS}^J|\beta) = \mathcal{L}_{r}(D_{\VS}^J|\beta) - \mathcal{L}_{c}(D_{\VS}^J), \\
    \label{eq:loss_vs1}&\mathcal{L}_r(D|\beta) =\frac{1}{\abs{J}}\sum_{j\in J}\sum_{(e,e',y) \in D^j_{\VS}}\frac{\mathcal{L}\left(e,e',y|\A_{\beta(j)}^{\omega,\theta}\right)}{\abs{D^j_{\VS}}}\\
    \label{eq:loss_vs2}&\mathcal{L}_c(D) =\min_{\substack{l \neq  l'\\\abs{C_l}>0\\|C_{l'}|>0}} \frac{1}{\abs{J}}\sum_{j\in J}\sum_{(e,e',\_) \in D^j_{\VS}}\frac{D(e,e'|\A_l^{\omega,\theta},\A_{l'}^{\omega,\theta})}{{\abs{D^j_{\VS}}}}
\end{align}
Our ``value system loss'' in Eq.~(\ref{eq:loss_vs}) has two terms. The first term, in Eq.~(\ref{eq:loss_vs1}) increments representativeness by minimizing discordance (Eq.~(\ref{eq:discordance})). The second term (Eq.~(\ref{eq:loss_vs2})) increases conciseness by separating the preference models of the most similar clusters. %Though the conciseness is mostly limited by the quality of the cluster assignment $\beta$, this penalty can aid the system to keep improving the conciseness of the subsequent assignments by forcing clustering separation. We use a penalty factor $\sigma>0$ to weight the penalty. In the experiments we found a best fit for $\sigma=1$. 
As conciseness is not differentiable, we employ a quantitative version of inter-cluster discordance (Eq.~(\ref{eq:discordance-inter})), the term $D(e,e'|A_1,A_2)$: the Jensen Shannon Divergence between the Bernoulli probability distributions of parameters $p(e,e'|A_1)$ and $p(e,e'|A_2)$. Incrementing this metric tends to increase $\delta(p(e,e'|\A^{\omega,\theta}_l),p(e,e'|\A^{\omega,\theta}_l))$, thus increasing conciseness. {Jensen-Shannon divergence has also been used in the related problem of finding the centroid of probability distributions~\cite{jensenfordistancebetweenprobscentroid}}.%We use a penalty factor $\sigma>0$ to weight this penalty term. In the experiments we found a best fit for $\sigma=1$.

%To do this, we multiply the part of the loss corresponding to each cluster $l$ by a proportional factor $r_l = 1-\repr_{\cup_{j\in C_l}D_{\VS}^j}\left(\preccurlyeq_{A_l \circ G^\theta_V}|C_l\right)$ --where $\preccurlyeq_{A_j \circ G^\theta_V}$ is the preorder relation induced by the function $A_j \circ G^\theta_V$-- is the negated representativeness of cluster $l$ according to our current models $A^\omega_l\circ G^\theta_V$. The second term of the loss, in Eq.~\ref{eq:loss_vs2}, is a penalty term weighted by a penalty factor $\sigma \geq 0$. It penalizes solutions with low discordance between clsuters, i.e., tries to increase conciseness by separating the preference models of the current clusters. Though the conciseness is mostly limited by the quality of the cluster assignment $\beta$, we hypothesize that this factor can aid the system to keep improving the conciseness of the subsequent assignments by forcing clustering separation to a certain extent. 

%In a similar fashion to the first term of the loss, we define for each compared pair of clusters the weighting factor $c_l^{l'} = 1-d_{D_{\VS}^{J}}(\preccurlyeq_{A_l^\omega \circ G_V^\theta},\preccurlyeq_{A_{l'}^\omega\circ G_V^\theta})$, that is the negated discordance between the value systems $A_l$ and $A_{l'}$. The weights $c_l$ penalize the clusters that are actually less discordant, which leads to a more directed increase in conciseness (given it is defined as the minimum over these discordances). 

The grounding loss for each value $v_i$, with $i=1,\dots, m$, (Eq.~(\ref{eq:loss_gr})) is a cross-entropy loss computed over its corresponding dataset $D_{v_i}$, aggregating the examples of each agent separately. Minimizing these losses increases grounding coherence by reducing discordances.  

\begin{equation}\label{eq:loss_gr}
    \mathcal{L}_{V}(D_{V}) = \left(\frac{1}{\abs{J}}\sum_{j\in J}\sum_{(e,e',y) \in D^j_{v_i}} \frac{\mathcal{L}(e,e',y)|\A_{v_i}^\theta)}{\abs{D^j_{v_i}}}\right)_{i=1}^m
\end{equation}

The grounding and value system loss functions need to be minimized in a hierarchy, i.e., prioritizing the grounding loss to improve coherence, and in second place, consider the value system loss. We approach this as a constrained optimization problem. The constraints to satisfy here are maximizing the coherence with each value, i.e., finding grounding network parameters $\theta$ such that $ \chr_{D_{v_i}}(\A^{\theta}_{v_i}) = \chr^*_{v_i}$, with $\chr^*_{v_i} = \max_{\theta \in \Theta}\chr_{D_{v_i}}(\A^{\theta}_{v_i})$, for every $i\in \{1,\dots, m\}$. Since $\chr^*_{v_i}$ is unknown a priori, it is dynamically estimated as the highest coherence observed during the learning process. The constraint to satisfy in terms of our loss function should be $\mathcal{L}_V(D_V) \leq \mathcal{L}^*_{V}$, where $\mathcal{L}^*_V$ is a loss that guarantees maximum coherence with all values. As we do not know $\mathcal{L}_V^*$, we assume the stricter constraint $\mathcal{L}_V(D_V) = 0$. With $m$ positive Lagrange multipliers $\lambda = (\lambda^1, \dots, \lambda^m)$ our objective is transformed to:

\begin{align}\label{eq:Lagrange}
    \min_{\theta,\omega} \max_\lambda\mathcal{L}_{\VS}(D_{\VS}^J|\beta) &- \lambda \cdot \left(\mathcal{L}_V(D_V)\right)^T
\end{align}

We seek a Nash equilibrium of jointly minimizing the Lagrangian in Eq.~(\ref{eq:Lagrange}) over $\theta,\omega$ (subject to the assignment $\beta$) and maximizing over $\lambda \in \R^+$~\cite{pmlrv98cotter19aLagrangianNash}. This is done through successive iterations of improving the Lagrangian (via gradient descent, Line 7) and then increasing the Lagrange multipliers $\lambda$ through gradient ascent with a learning rate $\alpha_\lambda$ (Line 9). To avoid overfitting the artificial constraint $\mathcal{L}_V(D_V) = 0$, the Lagrange multipliers for each value $v_i$ increase only when the coherence is below $\chr^*_{v_i}$.  Furthermore, multipliers are decayed using a factor $\gamma_\lambda$ if coherence remains at $\chr^*_{v_i}$.%Through this process, we optimize for the harder constraint $\mathcal{L}_V(D_V) = 0$, which is not strictly necessary to achieve a totally coherent grounding. To not overfit the constraint, we only increment the multipliers if the coherence is behind the optimal one $\chr^*_V$. As we do not initially know the values of $\chr^*_V$, we estimate $\chr^*_V$ as the maximum coherence found in the optimization loop. To facilitate the minimization of the value system loss, we decrement the multipliers using a decay factor $\gamma_\lambda$ if the coherence remains at $\chr^*_V$.  %The only assumption for this algorithm to work is that $\lambda_0$ must be big enough to increase the initial coherency.

%The resulting algorithm from joining the E-step and M-steps is depicted in Algorithm~\ref{alg:algorithm1}. 

\begin{algorithm}[t]
\caption{Value system learning of a society (EM algorithm)}\label{alg:algorithm1}
\textbf{Initialization:} Datasets $D_{\VS}^J$, $D_V$. Learning rates $\alpha_\theta$, $\alpha_\omega$, $\alpha_\lambda$. Lagrange multiplier decay $\gamma_\lambda > 0$. Maximum number of clusters $L_{max}$. Number of M-Steps in the first epoch ($b_0$), and on subsequent ones ($b_r$). Set maximum achievable coherence $\chr^*_{v_i} = 0$ for all $i$.%, and minimum needed grounding loss $L_V^* = \infty$.

\textbf{Input (at a given step of Algorithm~\ref{alg:algorithm2}):} 
Assignment $\beta_1$ (optional), parameters of the value system weights $\omega_0$, parameters of the grounding network $\theta_0$, number of epochs $R$. Lagrange multiplier state $\lambda_0 = \left(\lambda^i_0\right)_{i=1}^m$ (optional, otherwise use initialization).

\textbf{Output:} An assignment of agents into clusters $\beta$, updated parameters $\theta_R$, $\omega_R$ and new Lagrange multipliers $\lambda_R$ .

 \begin{algorithmic}[1]
    \State Set $G_V^\theta$ and $W_l^{\omega}$ (for $l<L_{max}$) with params. $\theta_{0}$ and $\omega_{0}$, resp.
    \For{epoch $r=0,\dots,R-1$}
    \State \textproc{\textbf{E-Step}} (omit if $\beta_1$ is supplied and $r=0$):
     %\For{$j\in J$}
        %\State $$\beta_r(j) \gets \argmin_{l<L_{max}} \prod_{i=1}^\abs{D^j_{\VS}} \left(\abs{y^j_i - p(e_i,e_i'|A_l\circ G^\theta_V)}\right)$$
        \State $\beta_{r+1}(j) \gets \argmin_{l} d_{D_{\VS}^j}\left(\A_{l}^{\omega,\theta}\right)$ \Comment{Do for all $j\in J$}
    %\EndFor
%    \State $K_r \gets \#\{\beta_r(j)|j\in J\}$
    \State \textproc{\textbf{M-Step}} (Repeat $b_r$ times):
    
    \State $\mathcal{L}_{global} =  \mathcal{L}_{\VS}(D_{\VS}^J|\beta_r) + \lambda_r\cdot  \left(\mathcal{L}_V(D_V)\right)^T$
    \State $\theta_{r+1} \gets \theta_{r} - \alpha_\theta\nabla_\theta\mathcal{L}_{global}  $; $\omega_{r+1} \gets \omega_{r} - \alpha_\omega\nabla_\omega\mathcal{L}_{global}  $ 
    \If{$\chr^*_{v_i} > \chr_{D_{v_i}}(\A_{v_i}^\theta)$} \Comment{Do 8-11 for all $i$}
    \State $\lambda^i_{r+1} \gets (1-\gamma_\lambda)\lambda^i_{r} + \alpha_{\lambda} \left(\mathcal{L}_{V}(D_V)\right)_i$
    \EndIf
    
    \State $\chr^*_{v_i} \gets \max\left(\chr_{D_{v_i}}(\A_{v_i}^\theta\right), \chr^*_{v_i})$
    \EndFor
    \State \textbf{Return} $\beta_R$, $\omega_{R}$, $\theta_{R}$, $\lambda_R$
\end{algorithmic}
\end{algorithm}

%This is not a relevant issue in the reference publication~\cite{pmlr-v235-chakraborty24b}, as they depart from a previous ``good enough'' model (a non-socially aware LLM). In our case, we have no prior solution, though.
%EM algorithms are known to converge to local optima or stationary points~\cite{emlocalconvergence}, depending on initialization. To address this, Algorithm~\ref{alg:algorithm2} introduces an exploitation-exploration outer loop inspired by evolutionary algorithms, extending the EM procedure in Algorithm~\ref{alg:algorithm1}. A memory $M$ (which acts as the evolutionary algorithm \textit{population}) of social value systems is maintained. At each iteration, a solution is selected from $M$ based on its quality (Line 5), mutated with probability $\epsilon > 0$ (Line 7), and then refined with Algorithm~\ref{alg:algorithm1} (Line 9) during $R$ epochs where the first cycle directly performs the M-step over the mutated solution (Line 3 Algorithm~\ref{alg:algorithm1}). After that, it returns a new social value system.
EM algorithms are known to converge to local optima or stationary points~\cite{emlocalconvergence}, depending on initialization. To address this, Algorithm~\ref{alg:algorithm2} introduces an exploitation-exploration outer loop inspired by evolutionary algorithms (EA), extending the EM procedure in Algorithm~\ref{alg:algorithm1}. A memory $M$ (that acts as the EA \textit{population}) of social value systems is kept. At each iteration, a solution is selected from $M$ based on its quality (Line 5), mutated with probability $\epsilon > 0$ (Line 7), and then refined with Algorithm~\ref{alg:algorithm1} (Line 9) during $R$ epochs where the first cycle directly performs the M-step over the mutated solution (Line 3 Algorithm~\ref{alg:algorithm1}). Finally, it returns a new social value system.

The new solution is inserted in the memory (Line 10), replacing an existing one if it Pareto-dominates it. Pareto dominance is based on grounding coherence, number of clusters, conciseness, and representativeness. The memory has a capacity $N$, requiring an elimination protocol under overflow (Line 11). We seek a balance between keeping quality solutions --according to coherence, Dunn Index and Pareto dominance-- for exploitation, and maintaining varied clusterings for exploration. The eliminated solution is chosen as the worst in the following lexicographic order: (1) higher number of clusters, (2) number of identical agent-cluster mappings, (3) number of dominating solutions, (4) grounding coherence (5) Dunn Index. Solutions with the best coherence and Dunn Index are always preserved. %This strategy promotes exploration while retaining the best-performing solutions.

The selection step (Line 5) involves first, ordering the options by the outer optimization objective (Dunn Index) and then by the inner objective (grounding coherence). This order inversion is intentional, as coherence can in all cases be improved via the Lagrange multiplier method, while Dunn Index, and in particular, conciseness is best improved through exploration. Then, a solution is chosen with probability proportional to its rank (following Eq. (2) from~\cite{linearRankSelectionEq2}). 

The mutation step (Line 7) involves two tasks. First, it either removes a cluster --redistributing its agents randomly-- or adds a new cluster, populated by reassigning agents to it with a probability $p_m$. Second, it perturbs the parameters of both the grounding network and value system weights using Gaussian noise, following classical evolutionary strategies~\cite{fogelevolutionary}. The magnitude of perturbation is scaled by the coherence error for $ \theta$ and the Dunn Index error for $\omega$.

\begin{algorithm}[h]
\caption{Value System Learning of a society with exploration}\label{alg:algorithm2}
\textbf{Input:}. All the initialization parameters from Algorithm~\ref{alg:algorithm1}. Number of training steps $T$. Memory of candidate solutions size $N$. Epochs per training step, $R$. Mutation probability $\epsilon_0 < 1$, agent reassignment probability $p_m$, network parameter mutation scale $s_m$. Initial Lagrange multipliers $\lambda_0 =  \left(\lambda^i_0\right)_{i=1}^m$, $\lambda^i_0 > 0$.

\textbf{Output:} An assignment of agents into clusters $\beta$, and trained grounding networks $G_V^\theta$ and value system weights $\{W^\omega_{l}\}_{l=1}^{L}$.

 \begin{algorithmic}[1]
    \State Initialize Algorithm~\ref{alg:algorithm1}
    \State Generate value system network parameters $\omega_0$, one for each $W_{l}^\omega$, and grounding parameters $\theta_0$;
    \State Repeat Line 2 $N$ times to fill memory $M$ (add multipliers $\lambda_0$).
    \For{training step $t = 0,\dots, T-1$}
    \State $\beta_t,\theta_t,\omega_t,\lambda_t \gets$\textproc{SelectSolution}($M$)
    \If{Rand$() < \epsilon_t$}
    \State $\beta_t,\theta_t,\omega_t$  $\gets$ \textproc{MutateSolution}($M$, $p_m$, $s_m$)
    \EndIf
    \State $\beta'_{t},\theta'_t,\omega'_{t} \lambda'_t\gets$\textproc{\textbf{Algorithm~\ref{alg:algorithm1}}}($\beta_t, \theta_t,\omega_t,R,\lambda_t$)    %\Comment{If mutation made, the first E-Step is omitted.}
    \State \textproc{InsertInMemory($\beta'_t,\theta'_t,\omega'_t,\lambda'_t$, $M$)}
    \State If $M$ is full: \textproc{EliminateWorstSolution(M)}
    \EndFor
     
 \State $\beta$, $G_V^\theta$, $\{W^\omega_{l}\}_{l=1}^{L}$$\gets$ \textproc{GetBestSolution(M)} 
 \State \textbf{return} $\beta_t$, $G_V^\theta$, $\{W^\omega_{l}\}_{l=1}^{L}$
 \end{algorithmic}
\end{algorithm}

\section{Evaluation}\label{sec:evaluation}

We analyse a real-world train route choice dataset from Switzerland~\cite{apollochoicepublication}, where 388 agents stated their preferred route among two options, 9 instances per agent (3,492 in total). Each route is characterized by 4 attributes: travel time, cost, number of interchanges, and headway time. Additionally, 6 agent-specific \textit{context features} were collected in the dataset: household income (dollars), car availability (boolean), and trip intentions: commuting, shopping, business and leisure (also boolean). The last four are exclusive. Context features are agent specific, e.g., they have the same values for all instances of an agent. In our formalism, each route is an entity in the train choice domain, the society $J$ comprises the 388 agents, and their pairwise preferences form the dataset $D_{\VS}^J$. We solely assume that if agent $j$ prefers route $r_i$ over $r'_i$, then $r_i \succ^j r'_i$ (i.e., $y_i^j = 1$), and vice versa.

We assume that the route choices were guided by three values: time efficiency, cost efficiency, and comfort.%~\footnote{Rooted in utilitarianism, we conceive efficiency as a technical value that correlates negatively with cost and positively with pleasure (comfort).} 
While the groundings for time and cost efficiency are based on travel time and cost, respectively, we presume that comfort depends on headway and interchanges: if a route has both lower headway and fewer interchanges, we consider it more comfortable. In cases where only one of the features is better, we assume no preference and let the model estimate comfort alignment freely. We construct the grounding dataset $D_V$ by comparing all the choice instances from the original dataset, but in terms of each of the previous value definitions.%, for again 9 comparisons per value and agent.

In our experiments, the grounding network $G_V^\theta$ is unaware of these value groundings and learns to replicate the preferences in $D_V$ using only the 4 route features of time, cost, headway and interchanges. $G_V^\theta$ is composed by $3$ neural networks (one per value) with 3 hidden layers (sizes 16–24–16) with \textit{Tanh} activations, followed by a negative \textit{softplus} output activation function. 
%This ensures that the value alignment estimation is a positive number, so it behaves as another cost-like feature like time or number of interchanges. 
The input features are preprocessed by scaling them in  $[0,1]$. The value system weights $\{W_l^\omega\}_{l=1}^{L_{max}}$ have parameters $\omega \in \R^3$, and are treated as in Section~\ref{sec:algorithms}.

We performed two experiments: first, we ran Algorithm~\ref{alg:algorithm1} with $L_{max}=1$ to evaluate the necessity of clustering; second, we ran Algorithm~\ref{alg:algorithm2} with an increasing number of clusters $L_{max} \in \{2,3,4,5,6,9,12\}$, each with ten different seeds. %\footnote{Five seeds is a sufficient number of tests, as shown in the standard errors of the figures presented, and quantitatively in the the supplementary material}. 
Hyperparameter selection per size of $L_{max}$ is detailed in the supplementary material. In all cases, we assessed the quality of the learned grounding function $G_V^\theta$ in terms of grounding coherence. Furthermore, we analysed the learned social value systems quantitatively, examining the number of clusters, conciseness, and representativeness. For the best social value system configuration found, we examined the diversity of the value system weights and reflected on how the contextual feature values (not used during training) are distributed across the clusters, to assess whether they reflect interpretable choice patterns.

In Table~\ref{tab:tablek1} we provide the results of the first experiment (Algorithm~\ref{alg:algorithm1} with $L_{max}=1$).%obtained after aggregating the results for 5 seeds of running Algorithm~\ref{alg:algorithm1} with $L_{max}=1$. 
We obtain a single value system that represents the society with an 80.7\% in average, i.e., for each agent, their choices are represented by an 80.7\%. Additionally, we obtained total grounding coherence ($1$) in all seeds, meaning all the value alignment preferences were estimated properly. This shows that the Lagrange multiplier ascent mechanism correctly prioritized grounding coherence over value system representativeness --we include additional ablation studies in the supplementary material. Lastly, we observe the learned value system is based totally on comfort, possibly because the model took advantage of the 28\% of cases where we allowed it to predict anything --the instances where not simultaneously headway and interchanges were smaller or bigger in one of the routes.

\begin{table}[h]
    \centering
    \resizebox{0.98\columnwidth}{!}{\begin{tabular}{>{\RaggedLeft}p{2.65cm}>{\RaggedLeft}p{0.87cm}>{\RaggedLeft}p{1.08cm}>{\RaggedLeft}p{1.01cm}>{\RaggedLeft}p{1.14cm}}
        \toprule
        VS (Time, Cost, Comf) & Repr. & Chr Time & Chr Cost & Chr Comf \\
        \midrule
        (0, 0, 0.99) ± 0.001 & 0.807 ± 0.005 & 1.000 ± 0.0 & 1.000 ± 0.0 & 1.000 ± 0.0\\
        \bottomrule
        \end{tabular}}
    \caption{Results achieved for 10 seeds with $L_{max}=1$ cluster.}
    \label{tab:tablek1}
\end{table}

\begin{table*}[t] 
\centering 
\begin{tabular}{p{0.49cm}>{\RaggedRight}p{2.5cm}>{\RaggedLeft}p{0.45cm}>{\RaggedLeft}p{0.5cm}>{\RaggedLeft}p{0.5cm}>{\RaggedLeft}p{1.0cm}>{\RaggedLeft}p{1.04cm}| >{\RaggedLeft}p{0.95cm} >{\RaggedLeft}p{0.95cm} >{\RaggedLeft}p{0.95cm}>{\RaggedLeft}p{0.95cm}  >{\RaggedLeft}p{0.95cm}>{\RaggedLeft}p{0.95cm} } 
\toprule 
Cl. $l$ &VS (Time,Cost,Comf) & $\abs{C_l}$ & Repr. & Conc. & Dunn In. & Avg Chr. & Income & Car & Comm. & Shopping & Business & Leisure \\ 
\midrule 
1 & (0.02, 0.05 , \textbf{0.92})    ±(0.03, 0.08 0.11) & 262.3 ±13.0 & 0.865 ±0.01 & - & - & - & 75090.1 ({$-$1.8\%}) & 0.37 ({$-$2.5\%}) & 0.31 ({+7.6\%}) & 0.09 ({+8.8\%}) & 0.06 ($-$39.2\%) & 0.55 ({+1.4\%})\\ 
2 & (\textbf{0.70}, 0.04, 0.26) ±(0.16, 0.03, 0.14) & 87.9 ±14.0 & 0.797 ±0.01 & -  & - & - & 84127.7 ({+9.9\%}) & 0.43 ({+15.3\%}) & 0.25 ({$-$11.5\%}) & 0.03 ({$-$62.5\%}) & 0.23 ({+\textbf{142.7}\%}) & 0.49 ({-8.9\%})\\
3 & (0.05, \textbf{0.89}, 0.059) ±(0.05, 0.08, 0.06) & 37.8 ±1.6 & 0.816 ±0.012 & - & - & - & 69293.8 ({$-$9.4\%}) & 0.31 ({$-$17.9\%}) & 0.20 ({$-$29.0\%}) & 0.16 ({\textbf{+92.3}\%}) & 0.03 ({$-$68.7\%}) & 0.61 ({+13.1\%})\\
\midrule 
Total & - & 388 & \textbf{0.845} ±0.007 & 0.429 ±0.025 & 2.770 ±0.165  & 1.000 ±0.0& 76507.73 & 0.38 & 0.29 & 0.08 & 0.09 & 0.54 \\ 
\bottomrule 
\end{tabular}
\caption{Left side: Average results (with standard deviation) over 10 seeds with $L_{max}=3$: cluster value system, number of agents and representativeness; and in the last row, the final representativeness, conciseness, Dunn Index and coherence. Right side: cluster averages and proportional deviations from the global feature average (last row) of the six context features. The last five features are binary, values indicate the proportion of agents reporting each feature.} 
\label{tab:merged_l3} 
\end{table*}

%1 & 75090.1 ({-1.85\%}) & 0.37 ({-2.53\%}) & 0.31 ({+7.61\%}) & 0.09 ({+8.79\%}) & 0.06 ({-39.21\%}) & 0.55 ({+1.36\%})\\
 %2 & 84127.7 ({+9.96\%}) & 0.43 ({+15.31\%}) & 0.25 ({-11.51\%}) & 0.03 ({-62.56\%}) & 0.23 ({+\textbf{142}.7\%}) & 0.49 ({-8.89\%})\\
 %3 & 69293.8 ({-9.43\%}) & 0.31 ({-17.86\%}) & 0.20 ({-29.04\%}) & 0.16 ({+92.31\%}) & 0.03 ({-68.74\%}) & 0.61 ({+13.13\%})\\

\begin{figure}
    \centering
    \includegraphics[width=0.72\linewidth,trim=0.4cm 0.5cm 0.2cm 0.95cm,clip]{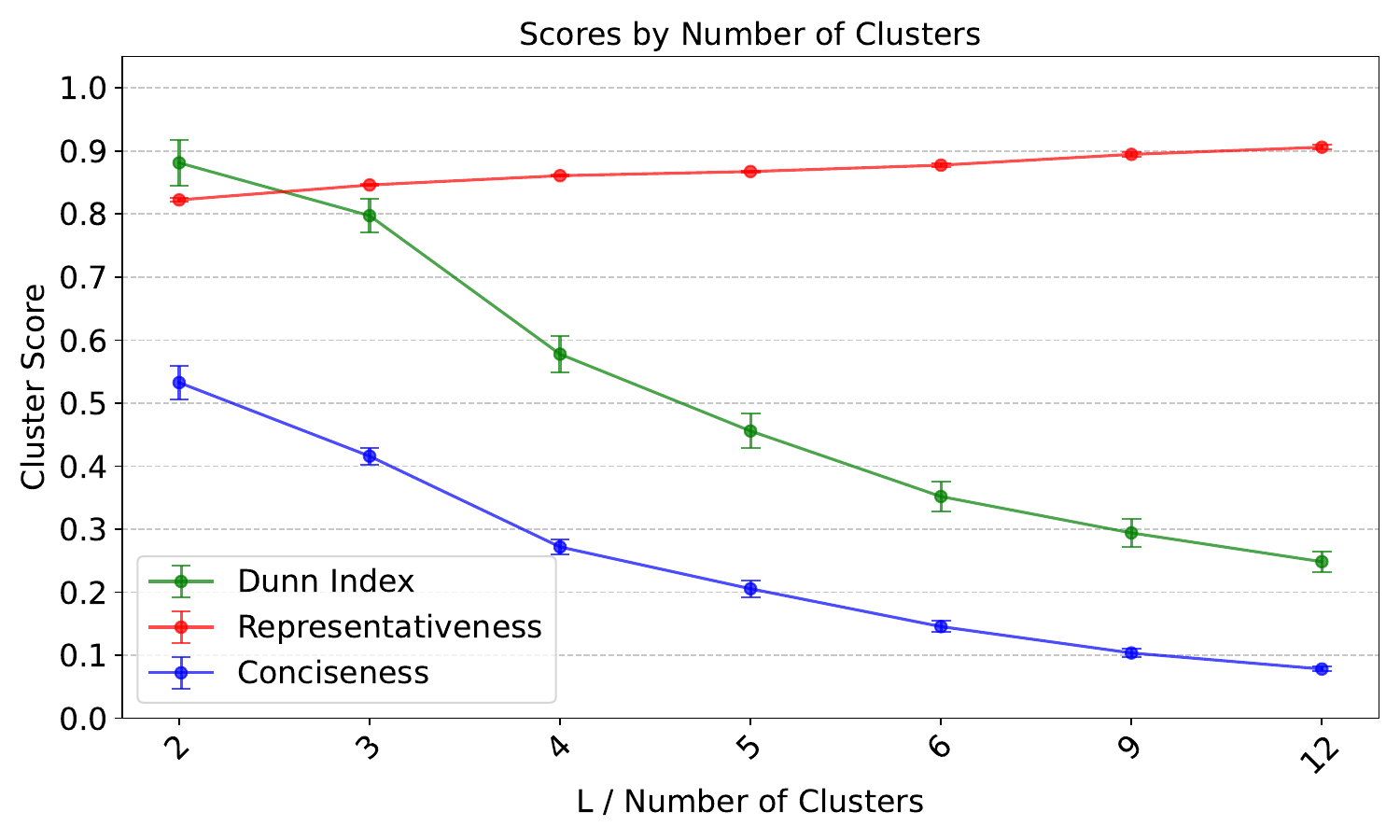}
    \caption{Normalized Dunn Index (scaled down by the maximum found), representativeness, and conciseness for experiments with $L_{max}$ ranging from 2 to 12. Each point shows the average and standard error over 10 seeds.}
    \label{fig:dunnindex}
\end{figure}

Figure~\ref{fig:dunnindex} shows value system scores across the tested $L_{max}$ values. The results follow a consistent trend: increasing $L_{max}$ improves representativeness but reduces conciseness. However, the number of clusters ($L$) found always matched $L_{max}$, reflecting a known limitation of the EM procedure, which favours representativeness over conciseness due to the greedy agent assignment step in Line 4, Algorithm~\ref{alg:algorithm1}. The best Dunn Index is achieved with $L_{max}=2$ with a representativeness of 0.815 in average. Note that this solution does not significantly improve representativeness compared to the $L_{max}=1$ solution. Thus, we consider the best configuration is achieved with $L=3$ clusters, where the representativeness advantage is more noticeable (84.5\%) while conciseness remains at a high level.

\begin{figure}
    \centering
    \includegraphics[width=0.72\linewidth,trim=0.41cm 0.5cm 0.3cm 0.9cm,clip]{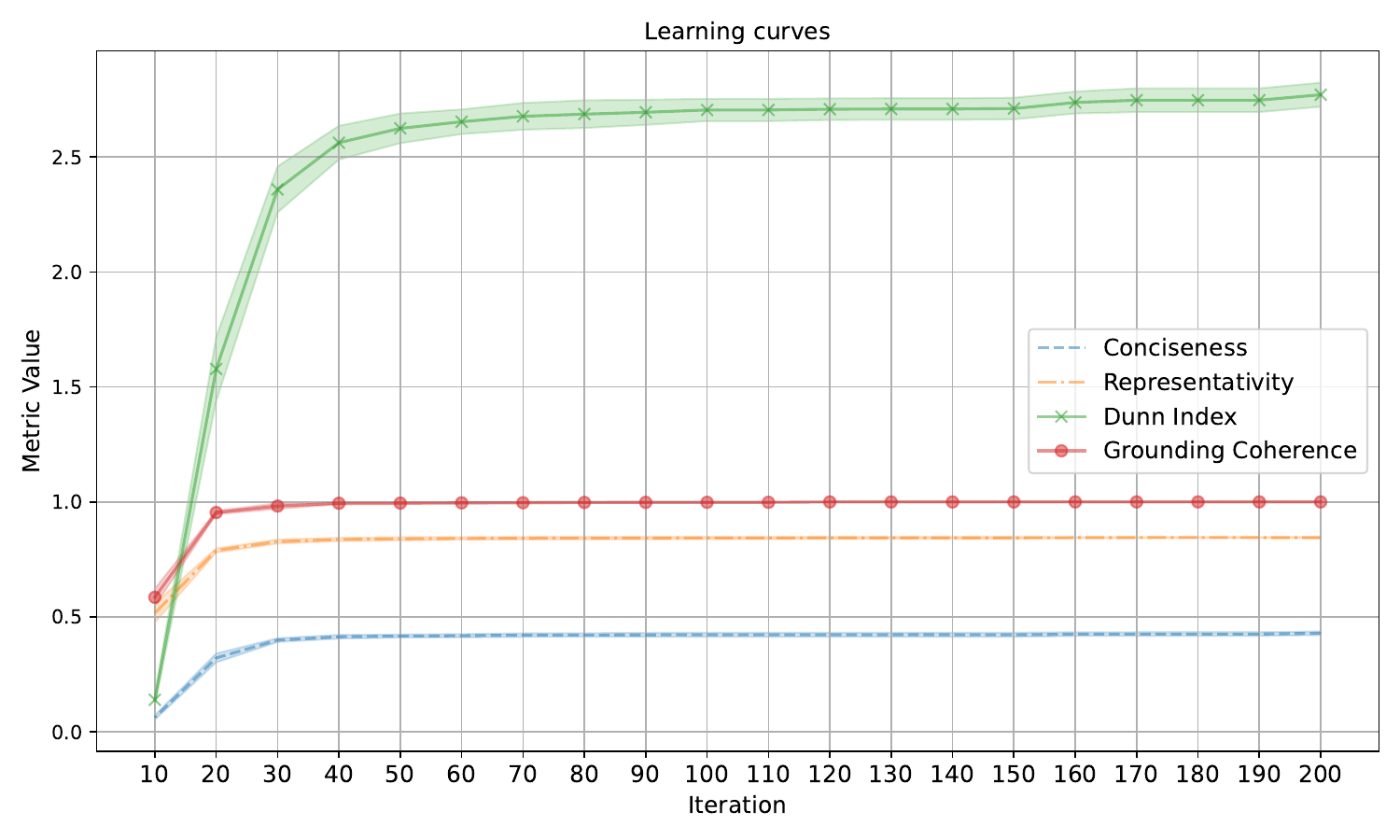}
    \caption{Learning curves for Dunn Index, representativeness, conciseness and grounding coherence of the best found clustering at each iteration (in terms of, first, coherence, and then Dunn Index): averages and standard errors from ten experiments ran with $L_{max}=3$ and different seeds.}
    \label{fig:learningcurve}
\end{figure}

In Figure~\ref{fig:learningcurve}, we present the aggregated learning curve for the selected case $L_{max}=3$, showing the mean and standard error across ten seeds. Notably, coherence rapidly reaches and maintains its maximum value (1) across all runs, empirically validating, again, the effectiveness of the Lagrange multiplier method. Representativeness and conciseness also improve steadily until a saturation point, beyond which further gains depend on occasional mutations.
%\todo[inline]{aunque sean 5 desviacion pequeña (supp material)}

Table~\ref{tab:merged_l3} shows the results achieved at the end of the learning process with $L_{max}=3$, averaging over ten seeds. Most agents ($\sim$262) were assigned to a comfort-based value system. Notably, this cluster's representativeness is 86.5\%, outperforming the single-cluster case and suggesting that some agents may be better represented by other values. The second-largest cluster ($\sim$88 agents) conveys a mix of comfort and time efficiency (26\% and 70\%, respectively), while the smallest group prioritizes cost (>89\%). Both smaller clusters achieve around 80\% representativeness, but the overall one improves over the $L=1$ case, reaching 85\%. The conciseness value indicates well-separated value systems --46.1\% of the preferences expressed by one value system cannot be represented by the others--.

%Most agents (around $240$) were assigned to a comfort-based value system. However, in this case the representativeness in this cluster surpassess the single cluster one, at over 86\%, indicating that other agents might as well be better represented by other values. The second biggest cluster with 77 agents was assigned a value system that combines comfort with time efficiency (26\% and 70\%, respectively), and the smaller group seem to mostly value cost (>97\%) over the other factors. In both cases we obtain a moderate representativeness of around 80\%, which at least is as good as if they were in a single cluster. Overall, the representativeness of the social value system yields higher, as expected, with about an 85\%. Finally, the conciseness is very similar among the clusters, meaning the value systems are very much separated (their models yield at \textit{minimum} an 46\% of times different preferences among the entities). 

We finish with a qualitative analysis for this case, in the right side of Table~\ref{tab:merged_l3}. We analyse the per-cluster distribution of context features and their relative change from the global averages (in percentages). In Cluster 1 there are no significant variations, except that it tends to include agents not on business trips. These are mostly included in Cluster 2 (+142.7\% business cases than the average). On the contrary, Cluster 3 gathers agents with shopping intentions (+92.3\% over average). Our algorithm reflected this pattern consistently across seeds. According to the model, for business trips, agents typically prioritize time efficiency, while for shopping, they prefer cheaper options, which likely to corresponds to reality. Also, agents in Cluster 3 tend to have less income or car availability, justifying their cost concerns.

\section{Conclusions and future work}\label{sec:conclusion}

In this paper we propose a formalization and a solution approach for the problem of learning explicit computational representations of the value system of a society of agents. In line with findings from social sciences, we acknowledge that different value systems co-exist in the same society. Setting out from a set of value labels, we learn a socially-derived computational semantics (value grounding functions) together with a set of value systems that represents the society's preference diversity while remaining concise. %The methodology proposed is funded on the social sciences and humanities literature, which suggests that a simple aggregation of value systems into a single one might misrepresent social groups.
We illustrate the real-world applicability of the approach in a use case on train trip choices, where decisions are guided by values such as time/cost efficiency and comfort. Groups of agents were assigned to a value system that not only represented their stated preferences, but also reflected their travel intentionality (e.g., for shopping, business). 

There are, of course, limitations to our work. %both regarding the societal value system model and the associated learning problem, as well as the approximation algorithm. 
As we argue in this paper, in general it seems reasonable to assume a socially-agreed grounding of values within a society, but in certain cases (e.g., in multi-cultural societies) this assumption may not hold. Furthermore, while our adaptation of the Dunn Index used to define a desired trade-off between conciseness and representativeness seems an obvious choice, it needs to be further supported by experimental studies. Limitations of the proposed heuristic approach include the difficulty in finding concise solutions in terms of the number of clusters and difficulties in interpreting the learned value grounding functions. 
%\textcolor{blue}{falta mencionar más limitaciones (intentar que table 2 aparezca en página 7, }

%As future work, we suggest making explicit the changes in agents' value systems produced by varying contexts.
As future work, we suggest making value systems adaptable to varying contexts.
Also, we propose making the algorithm adaptable to other analysis intentions by exploring alternative optimization metrics. Another interesting avenue for research is generalizing the approach to sequential DM, as well as to exploring learning agent-dependent value semantics and separating goal/task identification from value preferences. Analyzing the generalizability of the learned functions across the DM environment from limited datasets would be needed in those scenarios. Finding ways to represent the connection between agents' values and social values by drawing insights from sociology and cultural studies would also be fruitful.

\begin{ack}
This work is supported by grant VAE: TED2021-131295B-C33 funded by MCIN/AEI/10.13039/501100011033 and by ``European Union NextGenerationEU/PRTR'', by grant COSASS: PID2021-123673OB-C32 funded by MCIN/AEI/10.13039/501100011033 and by ``ERDF A way of making Europe'', and by the AGROBOTS Project of Universidad Rey Juan Carlos funded by the Community of Madrid, Spain.
\end{ack}

%%%%%%%%%%%%%%%%%%%%%%%%%%%%%%%%%%%%%%%%%%%%%%%%%%%%%%%%%%%%%%%%%%%%%%%%

%%% Use this command to include your bibliography file.

\bibliography{mybibfile}
\newpage
\mbox{}
\newpage

\section*{Supplementary Material for: \textit{Learning the Value Systems of Societies from Preferences} (ECAI 2025 paper id: M6755)}

\section*{Source Code}
Source code is available in the following Github repository \verb|https://github.com/andresh26-uam/| \verb|ValueLearningFromPreferences|.

\section*{Additional theoretical considerations}

\subsection*{On the bi-level optimization formulation}
At the end of Section~4 we claim that ``the bi-level optimization setup is needed instead of first estimating a coherent grounding and then trying to learn a social value system''. We prove this with a small counterexample where we can find a value system function that perfectly represents the value system preferences of an agent with a certain totally coherent grounding function but not with another one (that we could have learned without taking into consideration the agent's value system preferences).

Let two values \(v_1, v_2\) and three entities \(e_1, e_2, e_3\). Suppose one agent ($j$) reports that \(e_1 \succ_{v_1} e_3 \succ_{v_1} e_2\),  \(e_2 \succ_{v_2} e_3 \succ_{v_2} e_1\). A coherent grounding function $G_1$ could be: \(G_1(e_1) = (1,0)\), \(G_1(e_2) = (0,1)\), and \(G_1(e_3) = (0.3, 0.3)\). The agent also reports \(e_3 \prec^j_V e_2 \prec^j_V e_1\). On top of $G$, a totally representative value system function can be given through the weights \(w_1 = 0.6, w_2 = 0.4\). This shows that $G_1$ is a coherent grounding function that can be used to solve the bi-level optimization. Consider, instead, another coherent grounding $G_2$ with \(G_2(e_1) = (1,0)\), \(G_2(e_2) = (0,1)\), \(G_2(e_3) = (0.3, 0.7)\). In this case, since \(e_1 \succ^j e_2\), we require \(w_1 > w_2\), and since \(e_2 \succ^j e_3\), we need \(w_2 > 0.3w_1 + 0.7w_2\), which implies \(w_1 < w_2\) --a contradiction. Thus, no (linear) value system function can be found for $G_2$, despite it is also totally coherent. 

\subsection*{Alternative clustering metrics}

In Section 4, we proposed the Dunn Index as a clustering metric and optimization goal for the social value system learning problem. However, other metrics from the clustering literature could be easily adaptable to the characteristics of our setting. These include any metric that does not rely on calculating distances between cluster members, but only centroid-to-member or centroid-to-centroid distances. 
This is because, in our setting, cluster members are agents, and each of them compares different pairs of entities both with regard to value alignment ($D^j_V$) and value system ($D_{\textproc{VS}}^j$) preferences. On the other hand, the proposed distance is the discordance (Section~4, Equation~2), which relies on comparing the preferences over the same pairs of entities. When applied to a pair of agents, the proposed discordance metric would need to be applied over the pairs of entities that both agents have ranked (their intersection), which in most situations may be a small number of comparisons or even the empty set\footnote{The latter occurs in the presented dataset, as each agent labels different pairs of unique trips.}. This would yield an unfeasible or irrelevant discordance value between agent preferences. 
However, as the cluster centroids are defined by the preference relation represented by an utility model (i.e. an alignment model, based on a set of value system weights and a grounding function), this utility can be employed to rank the pairs of entities supplied by any individual cluster agent. This enables to calculate a discordance between the preference relation represented by the utility and that of any agent, measured over all the pairs of entities supplied by that particular agent. We use this to calculate/define representativeness, for example. 

Examples of metrics that are based solely on cluster centroid to centroid or cluster members to centroid distances are the Ray-Turi Index~\cite{rayTuri} and the Davies-Bouldin index~\cite{DaviesBouldin1979}. Further experiments using these clustering scores in different environments are left out of the scope of this paper, but certainly comprises another avenue for future work that we propose in the last section of the main paper.

\section*{Experimental Details}

\begin{table}[b]
    \centering
    \renewcommand{\arraystretch}{1.2}
    \setlength{\tabcolsep}{4pt}
    \resizebox{\columnwidth}{!}{\begin{tabular}{c|rrrrrrrr}
    \toprule
    $L_{\max}$ & 1 & 2 & 3 & 4 & 5 & 6 & 9 & 12 \\
    
    \midrule
    $\epsilon_0$ & 0.0 & 0.2 & 0.2 & 0.25 & 0.3 & 0.3 & 0.3 & 0.4 \\    
    $\lambda_0$ & 0.01 & 0.01 & 0.01 & 0.01 & 0.01 & 0.01 & 0.01 & 0.01 \\
    $\alpha_\lambda$ & 0.005 & 0.005 & 0.005 & 0.005 & 0.005 & 0.005 & 0.005 & 0.005 \\
    $\gamma_\lambda$ & $10^{-4}$ & $10^{-4}$ & $10^{-4}$ & $10^{-4}$ & $10^{-4}$ & $10^{-4}$ & $10^{-4}$ & $10^{-4}$ \\
    $\alpha_\theta$ & 0.005 & 0.005 & 0.005 & 0.005 & 0.005 & 0.005 & 0.006 & 0.006 \\
    $\alpha_\omega$ & 0.01 & 0.01 & 0.015 & 0.02 & 0.02 & 0.02 & 0.02 & 0.025 \\
    $T$ & 1 & 150 & 200 & 200 & 225 & 250 & 400 & 400 \\
    $N$ & - & 4 & 5 & 5 & 5 & 6 & 7 & 8 \\
    $R$ & 500 & 3 & 3 & 3 & 4 & 4 & 4 & 4 \\
    $b_r$ & 10 & 3 & 3 & 4 & 3 & 3 & 5 & 5 \\
    $b_0$ & 10 & 10 & 12 & 12 & 12 & 12 & 16 & 20 \\
    $p_m$ & 0 & 0.1 & 0.1 & 0.1 & 0.1 & 0.1 & 0.1 & 0.1 \\
    $s_m$ & 0 & 0.3 & 0.25 & 0.25 & 0.25 & 0.2 & 0.1 & 0.1 \\
    \bottomrule
    \end{tabular}}
    \caption{Hyperparameters used in each experiment for varying $L$.}
    \label{tab:hyperparameters}
\end{table}
In Table~\ref{tab:hyperparameters}, we include a comprehensive table of hyperparameters used in our experiments. The general rule we experimentally tested for the selection of parameters is that, for higher values of $L_{max}$, increasing memory size, learning rates, and iterations had a positive effect. On the contrary, mutation scale decreases as $L_{max}$ increases to favour the exploitation of existing solutions that are increasingly complex to optimize. The case $L_{max}=1$ was run only with Algorithm~1 (as it did not need exploration given there is only one possible assignment of agents into the single cluster) but, to avoid any bias, it was run during 500 epochs with 10 M-Steps repetitions in each epoch, which resulted in far more optimization steps (5000) than it was achieved with any of the other solutions with bigger $L_{max}$. This is due to the fact that, due to the probabilistic selection procedure, in the memory each solution was chosen for optimization and mutation only a handful of times per iteration of Algorithm~2. In particular, it was noted that the actual number of optimization steps for any particular candidate solution capped (experimentally) at around 1000 steps. We ran the experiments with 10 seeds (from number 26 to 35). 

\begin{table}[h]
    \centering
    \begin{tabular}{lp{6.5cm}}
        \toprule
        \textbf{Symbol} & \textbf{Description} \\
        \midrule
        $L_{\max}$ & Maximum number of clusters/components \\
        $\epsilon_0$ & Initial mutation probability (probability of mutating a solution) \\
        $\lambda_0$ & Initial Lagrange multipliers \\
        $\alpha_\lambda$ & Learning rate for Lagrange multipliers \\
        $\gamma_\lambda$ & Decay rate for Lagrange multipliers \\
        $\alpha_\theta$ & Learning rate for the grounding model $G_V^\theta$ parameters $\theta$ \\
        $\alpha_\omega$ & Learning rate for the value system weights $W_l^{\omega}$ parameters $\omega$ \\
        $T$ & Number of training iterations \\
        $N$ & Size of the clustering candidate list/memory\\
        $R$ & Number of times to run the EM-algorithm at each iteration \\
        $b_0$ & Initial M-Step repetitions (after retrieving from memory) \\
        $b_r$ & Subsequent M-Step repetitions (after E-Steps) \\
        $p_m$ & Agent reassignment probability (of moving an agent to another cluster) \\
        $s_m$ & Mutation scale for network parameters \\
        \bottomrule
    \end{tabular}
    \caption{Glossary of hyperparameters used in the experiments.}
    \label{tab:hp-glossary}
\end{table}
%%%%%%%%%%%%%%%%%%%%%%%%%%%%%%%%%%%%%%%%%%%%%%%%%%%%%%%%%%%%%%%%%%%%%%%%
\textbf{Hardware and approx. wall clock times}. The experiments where executed on a MacBook Pro with 16GB RAM, chip Apple M2. The code is not optimized for efficiency, as this was not in the scope of the paper. As such, the current implementation for the longest experiments ($L=12$) took $5.05$ hours in average with minimal deviation across seeds (approximately $\pm$ 10 minutes). For reference, $L=1$ took approximately 1.6 hours and $L=2$, $ 1.41$ hours.

\section*{Additional results}
To further motivate the advantages of the proposed bi-level optimization method, we provide two more baseline experiments.

The first baseline is a simple reward learning method based on fitting the Bradley-Terry model with no consideration of human values (e.g. as in Section 2.2 in RLHF~\cite{christiano2023deeprlpreferences}, with none of the mentioned modifications) and based on the value system preferences of the whole society considered as a single agent, using exactly the same network architecture as the one used for the experiments. Though we obtained a value system representativeness over 0.964 $\pm$0.016, the grounding coherence was, naturally, inadmissibly low (below or around $0.5$ in all values). This result implies that, unfortunately, the utility functions of the agents are more complex to explain than with solely a linear weighting scheme over simple to understand values. The advantage of our method then, consists of reaching a balance between accuracy and value-aware explainability of agent preferences. 

The second baseline certifies the advantage in accuracy gained over a naïve sequential optimization version of the social value system learning problem for $L=1$ cluster. This consisted of maximizing first grounding coherence by fitting the grounding networks for each separately with the losses $(\mathcal{L}_{V})_i$ ($i=1,2,3$), and then with these networks as the assumed fixed grounding, fitting value system weights that maximize two value system loss $\mathcal{L}_{\VS}$. Each step was run for 20000 gradient descent steps each (far more than with the experiment with $L_{max}=1$, with 5000 steps) and repeated 10 times (with seeds 26 to 35). Naturally, as in our main experiments, we obtained total coherence ($1.0$ for all values), but in average, we obtained a lower value system representativeness (0.750$\pm 0.010$) than that of any of the clustered solutions, and less so than our solution with $L_{max} =1$. This result further proves that the bi-level formulation was necessary not only as a theoretical consideration (see first section of the appendix), but also in this experimental case.

\newpage
Finally, we executed two experiments without the Lagrange multiplier ascent method: keeping the initial multiplier penalty at $\lambda_0=(0.01,0.01,0.01)$ (for all values), and eliminating Lines 8-11 from Algorithm~1. In the first experiment we set $L_{max}=1$ (Table~\ref{tab:nonlagrangian1}), and, in the second, we set $L_{max}=3$ (Table~\ref{tab:nonlagrangian3}). In both we observe that representativeness is higher than in the paper results. However, the coherence reduction is noticeable in both cases, suggesting the model neglected representing groundings for representing value systems instead. This empirically proves the necessity of updating the Lagrange multipliers as suggested in our approach to properly solve the bi-level formulation proposed.
 \\
\begin{table}[t]
    \centering
    \begin{tabular}{lrrrr}
\toprule
Value System & Repr. & Chr Time & Chr Cost & Chr Comf \\
\midrule
0.154, 0.349, 0.497 & 0.895 & 0.642 & 0.899 & 0.949 \\
\end{tabular}
    \caption{Results without the multiplier ascent method and $L_{max}=1$.}
    \label{tab:nonlagrangian1}
\end{table}

\begin{table}[t]
    \centering
    \resizebox{\columnwidth}{!}{
    \begin{tabular}{p{0.1cm}>{\raggedright}p{1.94cm}>{\raggedleft}p{0.15cm}>{\raggedleft}p{0.3cm}>{\raggedleft}p{0.3cm}>{\raggedleft}p{0.3cm}>{\raggedleft}p{0.3cm}
    >{\raggedleft}p{0.3cm}
    >{\raggedleft\arraybackslash}p{0.35cm}}
\toprule
Cl. $l$ & VS (Time, Cost, Comf) & $|C_l|$ & Repr. & Conc. & Dunn Ind. & Chr. Time & Chr. Cost & Chr. Comf \\
\midrule
1 & (0.01, 0.50, 0.49)  & 166 & 0.886  & -  & -  & - & -  & - \\
2 & (0.12, 0.00, 0.88)  & 149 & 0.853 & - & - & -  & -  & - \\
3 & (0.01, 0.98, 0.01)& 73 & 0.839 & -  & - & - &- &-  \\
\midrule
Tot. & (0.05, 0.40, 0.55)  & 388 & 0.864  & 0.261  & 1.920  & 0.916  & 0.788  & 0.901\\
\bottomrule
\end{tabular}}
\caption{Results without the multiplier ascent method and $L_{max}=3$}
\label{tab:nonlagrangian3}
\end{table}

\iffalse

\begin{table}[t]
    \centering
    \resizebox{0.99\columnwidth}{!}{
    \begin{tabular}{p{0.2cm}>{\raggedright}p{1.95cm}>{\raggedleft}p{0.2cm}>{\raggedleft}p{0.3cm}>{\raggedleft}p{0.3cm}>{\raggedleft}p{0.65cm}>{\raggedleft\arraybackslash}p{0.3cm}
    >{\raggedleft\arraybackslash}p{0.3cm}
    >{\raggedleft\arraybackslash}p{0.35cm}}
\toprule
Cl. $l$ & V. S. & $|C_l|$ & Repr. & Conc. & Dunn Index & Chr. Eff & Chr. Cost & Chr. Comf \\
\midrule
1 & (0.01, 0.50, 0.49)  & 166 & 0.886  & 0.261  & 2.277  & 0.904  & 0.796  & 0.892  \\
2 & (0.12, 0.00, 0.88)  & 149 & 0.853  & 0.261  & 1.774  & 0.943  & 0.805  & 0.896  \\
3 & (0.01, 0.98, 0.01)& 73 & 0.839  & 0.286  & 1.771  & 0.887  & 0.738  & 0.932  \\
\midrule
Tot. & (0.05, 0.40, 0.55)  & 388 & 0.864  & 0.261  & 1.920  & 0.916  & 0.788  & 0.901  \\
\bottomrule
\end{tabular}}
\caption{Results obtained without the Lagrange multiplier ascent method and $L_{max}=3$}
\label{tab:nonlagrangian3}
\end{table}
\fi

\end{document}